\documentclass{article}

\usepackage[nonatbib, final]{neurips_2019}
\usepackage[
    natbib=true,
    style=numeric,
    maxnames=4,
    minnames=4,
    maxbibnames=99,
    backend=bibtex,
    urldate=iso8601,
    firstinits=true,
    isbn=false,
    doi=false,
    date=year,
    sorting=nty,
]{biblatex}
\DeclareFieldFormat
  [article,inbook,incollection,inproceedings,patent,phdthesis,thesis,unpublished]
  {title}{#1\isdot}  %
\usepackage{graphicx}
\graphicspath{{images/},{images-bamos-unorg/}}
\usepackage{caption}
\usepackage{subcaption}
\usepackage{wrapfig}
\usepackage{adjustbox}
\usepackage{mathtools}

\usepackage[utf8]{inputenc} %
\usepackage[T1]{fontenc}    %
\usepackage{color}
\definecolor{link_color}{RGB}{0,128,255}
\usepackage[%
colorlinks=true,allcolors=link_color,pageanchor=true,%
plainpages=false,pdfpagelabels,bookmarks,bookmarksnumbered,%
]{hyperref}
\usepackage{url}            %
\usepackage{booktabs}       %
\usepackage{amsfonts}       %
\usepackage{nicefrac}       %
\usepackage{microtype}      %

\usepackage{amsthm}
\newtheorem{lemma}{Lemma}

\usepackage{todonotes}

\usepackage{tikz}

\usepackage{xspace}
\newcommand{\qpth}{\texttt{qpth}\xspace}

\usepackage{listings,textcomp,color}
\definecolor{backcolour}{rgb}{0.96,0.96,0.96}
\definecolor{deepblue}{rgb}{0,0,0.5}
\definecolor{deepred}{rgb}{0.6,0,0}
\usepackage{amsmath}

\DeclareMathAlphabet\mathbfcal{OMS}{cmsy}{b}{n}

\newcommand{\DD}{\mathsf{D}}

\newcommand{\norm}[1]{\|{#1}\|}

\usepackage[capitalise,nameinlink,noabbrev]{cleveref}

\newcommand{\BEAS}{\begin{eqnarray*}}
\newcommand{\EEAS}{\end{eqnarray*}}
\newcommand{\BEA}{\begin{eqnarray}}
\newcommand{\EEA}{\end{eqnarray}}
\newcommand{\BEQ}{\begin{equation}}
\newcommand{\EEQ}{\end{equation}}
\newcommand{\BIT}{\begin{itemize}}
\newcommand{\EIT}{\end{itemize}}
\newcommand{\BNUM}{\begin{enumerate}}
\newcommand{\ENUM}{\end{enumerate}}

\newcommand{\BA}{\begin{array}}
\newcommand{\EA}{\end{array}}

\newcommand{\eg}{{\it e.g.}}
\newcommand{\ie}{{\it i.e.}}

\newcommand{\reals}{{\mbox{\bf R}}}

\usepackage{algorithm}
\usepackage{algorithmicx}
\usepackage{algpseudocode}
\newcounter{module}

\crefname{module}{Module}{Modules}

\title{Differentiable Convex Optimization Layers}

\author{%
  Akshay Agrawal \\
  Stanford University \\
  \texttt{akshayka@cs.stanford.edu} \\
  \And
  Brandon Amos \\
  Facebook AI \\
  \texttt{bda@fb.com} \\
  \And
  Shane Barratt \\
  Stanford University \\
  \texttt{sbarratt@stanford.edu} \\
  \AND
  Stephen Boyd \\
  Stanford University \\
  \texttt{boyd@stanford.edu} \\
  \And
  Steven Diamond \\
  Stanford University \\
  \texttt{diamond@cs.stanford.edu} \\
  \And
  J.~Zico Kolter\thanks{Authors listed in alphabetical order.}  \\
  Carnegie Mellon University \\
  Bosch Center for AI \\
  \texttt{zkolter@cs.cmu.edu}
}
\begin{document}

\maketitle

\begin{abstract}
Recent work has shown how to embed \emph{differentiable optimization problems}
(that is, problems whose solutions can be backpropagated through)
as layers within deep learning architectures. This method provides a useful
inductive bias for certain problems, but existing software for differentiable
optimization layers is rigid and difficult to apply to new settings. In this
paper, we propose an approach to differentiating through disciplined
convex programs, a subclass of convex optimization problems used by
domain-specific languages (DSLs) for convex optimization. We introduce
\emph{disciplined parametrized programming}, a subset of disciplined convex
programming, and we show that every disciplined parametrized program
can be represented as the composition of an affine map from parameters
to problem data, a solver, and an affine map from the solver’s solution to a
solution of the original problem (a new form we refer to as
\emph{affine-solver-affine} form). We then demonstrate how to efficiently
differentiate through each of these components, allowing for end-to-end
analytical differentiation through the entire convex program. We implement our
methodology in version 1.1 of CVXPY, a popular Python-embedded DSL for convex
optimization, and additionally implement differentiable layers for disciplined
convex programs in PyTorch and TensorFlow 2.0. Our implementation significantly
lowers the barrier to using convex optimization problems in differentiable
programs. We present applications in linear machine learning models and in
stochastic control, and we show that our layer is competitive (in execution
time) compared to specialized differentiable solvers from past work.
\end{abstract}

\section{Introduction}
\label{sec:intro}
Recent work has shown how to differentiate through specific subclasses of
convex optimization problems, which can be viewed as functions mapping problem
data to solutions \citep{amos2017optnet, djolonga2017differentiable,
  barratt2018differentiability, agrawal2019derivative, amos2019differentiable}.
These layers have found
several applications
\citep{geng2019coercing, amos2017optnet, donti2017task, de2018end,
amos2018differentiable, ling2018game, wilder2018melding, lee2019meta,
barratt2019least, barratt2019fitting}, but many applications remain relatively
unexplored (see, \eg, \citep[\S8]{amos2019differentiable}).

While convex optimization layers can provide useful inductive bias in
end-to-end models, their adoption has been slowed by how difficult they are
to use. Existing layers (\eg, \citep{amos2017optnet, agrawal2019derivative})
require users to transform their problems into rigid canonical forms by hand.
This process is tedious, error-prone, and
time-consuming, and often requires familiarity with convex analysis.
Domain-specific languages (DSLs) for convex optimization
abstract away the process of converting problems to canonical forms, letting
users specify problems in a natural syntax; programs are then lowered to
canonical forms and supplied to
numerical solvers behind-the-scenes \citep{agrawal2018rewriting}. DSLs
enable rapid prototyping and make convex optimization accessible to scientists
and engineers who are not necessarily experts in optimization.

The point of this paper is to do what DSLs have done
for convex optimization, but for differentiable convex optimization layers.
In this work, we show how to efficiently differentiate through disciplined
convex programs \citep{grant2006disciplined}.
This is a large class of convex optimization problems that can be
parsed and solved by most DSLs for convex optimization,
including CVX \citep{grant2014cvx}, CVXPY
\citep{diamond2016cvxpy, agrawal2018rewriting}, Convex.jl
\citep{udell2014cvxjl}, and CVXR \citep{fu2017cvxr}. Concretely, we introduce
\emph{disciplined parametrized programming} (DPP), a grammar for producing
parametrized disciplined convex programs. Given a program produced by DPP, we
show how to obtain an affine map from parameters to problem data, and an affine
map from a solution of the canonicalized problem to a solution of the original
problem. We refer to this representation of a problem --- \ie, the
composition of an affine map from parameters to problem data, a solver, and an
affine map to retrieve a solution ---  as \emph{affine-solver-affine} (ASA) form.

Our contributions are three-fold:

\textbf{1.} We introduce DPP, a new grammar for parametrized convex optimization
problems, and ASA form, which ensures that the mapping from problem parameters
to problem data is affine. DPP and ASA-form make it possible to differentiate
through DSLs for convex optimization, \emph{without explicitly backpropagating
through the operations of the canonicalizer}. We present DPP and ASA form in
\S\ref{sec:derivative}.

\textbf{2.} We implement the DPP grammar and a reduction from
parametrized programs to ASA form in CVXPY 1.1. We also implement
differentiable convex optimization layers in PyTorch
\citep{paszke2017automatic} and TensorFlow 2.0 \citep{agrawal2019tensorflow}.
Our software substantially lowers the barrier to using convex optimization
layers in differentiable programs and neural networks (\S\ref{sec:impl}).

\textbf{3.} We present applications to sensitivity analysis for linear machine
learning models, and to learning control-Lyapunov policies for stochastic control
(\S\ref{sec:ex}). We also show that for quadratic programs (QPs),
our layer's runtime is competitive with OptNet's specialized solver
\verb!qpth! \citep{amos2017optnet} (\S\ref{sec:eval}).

\section{Related work}
\paragraph{DSLs for convex optimization.}
DSLs for convex optimization allow users to specify convex optimization
problems in a natural way that follows the math. At the foundation of these
languages is a ruleset from convex analysis known as disciplined convex
programming (DCP) \citep{grant2006disciplined}. A mathematical program written
using DCP is called a disciplined convex program, and all such programs are
convex. Disciplined convex programs can be \emph{canonicalized} to
cone programs by expanding each nonlinear function into its graph
implementation \citep{grant2008graph}. DPP can be seen as a subset
of DCP that mildly restricts the way parameters (symbolic constants) can be used;
a similar grammar is described in \citep{chu2013code}. The techniques used in
this paper to canonicalize parametrized programs are similar to the methods
used by code generators for optimization problems, such as CVXGEN
\citep{mattingley2012cvxgen}, which targets QPs, and QCML, which targets
second-order cone programs (SOCPs)
\citep{chu2013code, chu2017qcml}.

\paragraph{Differentiation of optimization problems.} Convex optimization
problems do not in general admit closed-form solutions. It is nonetheless
possible to differentiate through convex optimization problems by implicitly
differentiating their optimality conditions (when certain regularity conditions
are satisfied) \citep{fiacco1968, robinson1980, amos2017optnet}. Recently,
methods were developed to differentiate through convex cone programs
in
\citep{busseti2018solution, agrawal2019derivative}
and
\citep[\S7.3]{amos2019differentiable}.
Because every convex program can be cast as a cone program,
these methods are general. The software released alongside
\citep{agrawal2019derivative}, however, requires users to express their
problems in conic form. Expressing a convex optimization problem in conic form
requires a working knowledge of convex analysis. Our work abstracts away conic
form, letting the user differentiate through high-level descriptions of convex
optimization problems; we canonicalize these descriptions to cone programs on the
user's behalf. This makes it possible to rapidly experiment with new families
of differentiable programs, induced by different kinds of convex optimization problems.

Because we differentiate through a cone program by implicitly differentiating
its solution map, our method can be paired with any algorithm for solving
convex cone programs. In contrast, methods that differentiate through every
step of an optimization procedure must be customized for each algorithm (\eg,
\citep{domke2012generic, diamond2017unrolled,
mardani2018neural}). Moreover, such methods only approximate the derivative,
whereas we compute it analytically (when it exists).

\section{Background}\label{sec:bg}

\paragraph{Convex optimization problems.}
A parametrized convex optimization
problem can be represented as
\begin{equation}\label{eqn-cvx}
\begin{array}{ll}
\mbox{minimize} & f_0(x; \theta) \\
\mbox{subject to} & f_i(x; \theta) \leq 0, \quad i=1, \ldots, m_1,\\
& g_i(x; \theta) = 0, \quad i=1, \ldots, m_2,
\end{array}
\end{equation}
where $x \in \reals^n$ is the optimization variable and
$\theta \in \reals^p$ is the parameter vector \citep[\S 4.2]{boyd2004convex}.
The functions $f_i : \reals^n \to \reals$ are
convex,
and the functions $g_i : \reals^n \to \reals$ are affine. A \emph{solution} to \eqref{eqn-cvx}
is any vector $x^\star \in \reals^n$ that minimizes the objective function,
among all choices that satisfy the constraints. The problem \eqref{eqn-cvx} can
be viewed as a (possibly multi-valued) function that maps a parameter to
solutions. In this paper, we consider the case when this solution map is
single-valued, and we denote it by $\mathcal S : \reals^p \to \reals^n$. The function
$S$ maps a parameter $\theta$ to a solution $x^\star$. From the perspective of
end-to-end learning, $\theta$ (or parameters it depends on) is learned in order
to minimize some scalar function of $x^\star$. In this paper, we show how to
obtain the derivative of $\mathcal S$ with respect to $\theta$, when
\eqref{eqn-cvx} is a DPP-compliant program (and when the derivative exists).

We focus on convex optimization because it is a powerful modeling tool, with
applications in
control \citep{boyd1994linear, bertsekas2005dynamic, todorov2012mujoco},
finance \citep{markowitz1952portfolio, boyd2017multi}, energy management
\citep{moehle2018dynamicnot}, supply chain
\cite{bertsimas2004robust,ben2005retailer}, physics \cite{kanno2011nonsmooth, angeris2019},
computational geometry \citep{van2000computational}, aeronautics
\citep{hoburg2014geometric}, and circuit design \citep{hershenson2001opamp,
boyd2005digital}, among other fields.

\paragraph{Disciplined convex programming.}
DCP is a grammar for constructing convex optimization problems
\citep{grant2006disciplined, grant2008graph}. It consists of functions,
or \emph{atoms}, and a single rule for composing them. An atom is a function
with known curvature (affine, convex, or concave) and per-argument
monotonicities. The composition rule is based on the following theorem from
convex analysis. Suppose $h : \reals^k \to \reals$ is convex, nondecreasing in
arguments indexed by a set $I_1 \subseteq \{1, 2, \ldots, k\}$, and
nonincreasing in arguments indexed by $I_2$. Suppose also that $g_i : \reals^n \to
\reals$ are convex for $i \in I_1$, concave for $i \in I_2$, and affine for $i \in
(I_1 \cap I_2)^c$. Then the composition $f(x) = h(g_1(x), g_2(x), \ldots,
g_k(x))$ is convex. DCP allows atoms to be composed so long as the composition
satisfies this composition theorem. Every disciplined convex program is a
convex optimization problem, but the converse is not true. This is not a
limitation in practice, because atom libraries are extensible (\ie, the class
corresponding to DCP is parametrized by which atoms are implemented). In this
paper, we consider problems of the form \eqref{eqn-cvx} in which the functions
$f_i$ and $g_i$ are constructed using DPP, a version of DCP that performs
parameter-dependent curvature analysis (see \S\ref{sec:dpp}).

\paragraph{Cone programs.}
A (convex) cone program is an optimization problem of the form
\begin{equation}
      \begin{array}{ll}
        \text{minimize}   & c^T x\\
        \text{subject to} &  b-Ax \in \mathcal K,\\
      \end{array}
      \label{eq:cp}
\end{equation}
where $x\in \reals^n$ is the variable (there are several other equivalent
forms for cone programs). The set $\mathcal{K}\subseteq \reals^m$
is a nonempty, closed, convex cone, and the
\emph{problem data} are $A \in \reals^{m \times n}$, $b\in \reals^m $, and
$c\in \reals^n$. In this paper we assume that \eqref{eq:cp} has a unique
solution. 

Our method for differentiating through disciplined convex programs requires
calling a solver (an algorithm for solving an optimization problem) in
the forward pass. We focus on the special case in which the solver is a
\emph{conic solver}. A conic solver targets convex cone
programs, implementing a function $s: \reals^{m \times n} \times \reals^m
\times \reals^n \to \reals^n$ mapping the problem data $(A,b,c)$ to a
solution $x^\star$.

DCP-based DSLs for convex optimization can canonicalize
disciplined convex programs to equivalent cone programs,
producing the problem data $A, b$, $c$, and $\mathcal K$
\citep{agrawal2018rewriting}; $(A, b, c)$ depend on the parameter $\theta$ and
the canonicalization procedure.  These data are
supplied to a conic solver to obtain a solution; there are many high-quality
implementations of conic solvers (\eg, \citep{odonoghue2017scs, mosek,
domahidi2013ecos}).

\section{Differentiating through disciplined convex programs}\label{sec:derivative}

We consider a disciplined convex program with variable $x \in
\reals^n$, parametrized by $\theta \in \reals^p$; its solution map can be viewed as a
function $\mathcal S : \reals^p \to \reals^n$ that maps parameters to the solution (see
\S\ref{sec:bg}). In this section we describe the form of $\mathcal S$ and how
to evaluate $\mathsf{D}^T \mathcal S$, allowing us to backpropagate through
parametrized disciplined convex programs. (We use the notation $\DD f(x)$ to
denote the derivative of a function $f$ evaluated at $x$, and $\DD^T f(x)$ to
denote the adjoint of the derivative at $x$.) We consider the special case of
canonicalizing a disciplined convex program to a cone program. With little
extra effort, our method can be extended to other targets.

We express $\mathcal S$ as the composition $R \circ s \circ C$; the
canonicalizer $C$ maps parameters to cone problem data $(A,b,c)$, the cone
solver $s$ solves the cone problem, furnishing a solution $\tilde x^\star$, and
the retriever $R$ maps $\tilde x^\star$ to a solution
$x^\star$ of the original problem. A problem is in ASA form if $C$ and $R$ are
affine.

By the chain rule, the adjoint of the derivative of a disciplined convex
program is
\[
\mathsf{D}^T\mathcal{S}(\theta) =
  \mathsf{D}^TC(\theta)
  \mathsf{D}^Ts(A,b,c)
  \mathsf{D}^TR(\tilde x^\star).
\]
The remainder of this section proceeds as follows. In \S\ref{sec:dpp}, we present
DPP, a ruleset for constructing disciplined convex programs reducible to ASA form. In
\S\ref{sec:canon}, we describe the canonicalization procedure and show how to
represent $C$ as a sparse matrix. In \S\ref{sec:solver}, we review how to
differentiate through cone programs, and in \S\ref{sec:retrieval}, we describe
the form of $R$.

\subsection{Disciplined parametrized programming}\label{sec:dpp}
DPP is a grammar for producing parametrized disciplined convex programs
from a set of functions, or atoms, with known curvature (constant, affine,
convex, or concave) and per-argument monotonicities. A program produced using
DPP is called a disciplined parametrized program. Like DCP, DPP is based on the
well-known composition theorem for convex functions, and it guarantees that
every function appearing in a disciplined parametrized program is affine,
convex, or concave. Unlike DCP, DPP also guarantees that the produced program
can be reduced to ASA form.

A disciplined parametrized program is an optimization problem of the form
\begin{equation}\label{eqn:dpp}
\begin{array}{ll}
  \mbox{minimize}   & f_0(x, \theta) \\
  \mbox{subject to} & f_i(x, \theta) \leq \tilde{f}_i(x, \theta), \quad i = 1, \ldots, m_1, \\
                    & g_i(x, \theta) = \tilde{g}_i(x, \theta), \quad i = 1, \ldots, m_2,
\end{array}
\end{equation}
where $x \in \reals^n$ is a variable, $\theta \in \reals^p$ is a parameter,
the $f_i$ are convex, $\tilde{f}_i$ are concave, $g_i$ and $\tilde{g}_i$ are
affine, and the expressions are constructed using DPP\@. An
expression can be thought of as a tree, where the
nodes are atoms and the leaves are variables, constants, or parameters. A
parameter is a symbolic constant with known properties such as sign but unknown
numeric value. An expression is said to be parameter-affine if it does not
have variables among its leaves and is affine in its parameters; an
expression is parameter-free if it is not parametrized, and variable-free
if it does not have variables.

Every DPP program is also DCP, but the converse is
not true. DPP generates programs reducible to ASA form by introducing
two restrictions on expressions involving parameters:
\begin{enumerate}
\item In
DCP, we classify the curvature of each subexpression appearing in the problem
description as convex, concave, affine, or constant. All parameters are
classified as constant. In DPP, parameters are classified as affine, just like
variables.
\item In DCP, the product atom $\phi_\mathrm{prod}(x, y) = xy$ is affine if $x$
or $y$ is a constant (\ie, variable-free). Under DPP, the
product is affine when at least one of the following is true:
\begin{itemize}
\item  $x$ or $y$ is constant (\ie, both parameter-free and variable-free);
\item one of the expressions is parameter-affine and the other is parameter-free.
\end{itemize}
\end{enumerate}

The DPP specification can (and may in the future) be extended to handle several
other combinations of expressions and parameters.

\paragraph{Example.} Consider the program
\begin{equation}\label{eqn:asa}
\begin{array}{ll}
  \mbox{minimize}  & \norm{Fx - g}_2 + \lambda \norm{x}_2 \\
  \mbox{subject to} & x \geq 0,
\end{array}
\end{equation}
with variable $x \in \reals^n$ and parameters $F \in \reals^{m \times n}$, $g \in \reals^m$,
and $\lambda > 0$. If $\norm{\cdot}_2$, the product, negation, and the sum are
atoms, then this problem is DPP-compliant:
\begin{itemize}
\item $\phi_\mathrm{prod}(F, x) = Fx$ is affine because
the atom is affine ($F$ is parameter-affine and $x$ is parameter-free) and $F$
and $x$ are affine;
\item $Fx - g$ is affine because $Fx$ and $-g$ are affine and the sum of affine
expressions is affine;
\item $\norm{Fx- g}_2$ is convex because $\norm{\cdot}_2$ is convex and convex
composed with affine is convex;
\item $\phi_{\mathrm{prod}}(\lambda, \norm{x}_2)$ is convex because the product
is affine ($\lambda$ is parameter-affine, $\norm{x}_2$ is parameter-free), it
is increasing in $\norm{x}_2$ (because $\lambda$ is nonnegative), and
$\norm{x}_2$ is convex;
\item the objective is convex because the sum of convex expressions is convex.
\end{itemize}

\paragraph{Non-DPP transformations of parameters.} It is often possible to
re-express non-DPP expressions in DPP-compliant ways. Consider the following
examples, in which the $p_i$ are parameters:
\begin{itemize}
\item The expression $\phi_{\mathrm{prod}}(p_1, p_2)$ is not DPP
because both of its arguments are parametrized. It can be rewritten in a
DPP-compliant way by introducing a
variable $s$, replacing $p_1p_2$ with the expression $p_1 s$, and adding the
constraint $s = p_2$.
\item Let $e$ be an expression.
The quotient $e/p_1$ is not DPP, but it can be rewritten as $ep_2$,
where $p_2$ is a new parameter representing $1/p_1$.
\item The expression $\log |p_1|$ is not DPP because
$\log$ is concave and increasing but $|\cdot|$ is convex. It can be
rewritten as $\log p_2$ where $p_2$ is a new parameter
representing $|p_1|$.
\item If $P_1 \in \reals^{n \times n}$ is a parameter representing a
(symmetric) positive semidefinite matrix and $x \in \reals^n$ is a variable,
the expression $\phi_{\mathrm{quadform}}(x, P_1) = x^T P_1 x$ is not DPP\@. It
can be rewritten as $\norm{P_2 x}_2^2$, where $P_2$ is a new parameter
representing $P_1^{1/2}$.
\end{itemize}

\subsection{Canonicalization}\label{sec:canon}
The canonicalization of a disciplined parametrized program to ASA form is
similar to the canonicalization of a disciplined convex program to a cone
program. All nonlinear atoms are expanded into their graph implementations
\citep{grant2008graph}, generating affine expressions of variables.
The resulting expressions are also affine in the problem parameters
due to the DPP rules. Because these expressions represent the problem data for the cone
program, the function $C$ from parameters to problem data is affine.

As an example, the DPP program \eqref{eqn:asa} can be canonicalized to
the cone program
\begin{equation}\label{eq:graph}
\begin{array}{ll}
  \mbox{minimize}  & t_1 + \lambda t_2 \\
  \mbox{subject to}  & (t_1, Fx - g) \in \mathcal{Q}_{m+1}, \\
                     & (t_2, x) \in \mathcal{Q}_{n+1}, \\
                     & x \in \reals^{n}_{+},
\end{array}
\end{equation}
where $(t_1, t_2, x)$ is the variable, $\mathcal{Q}_n$ is the
$n$-dimensional second-order cone, and $\reals^n_+$ is the nonnegative orthant.
When rewritten in the standard form \eqref{eq:cp}, this problem has data
\begin{align*}
A = \begin{bmatrix}
-1 &  &    \\
   &  & -F  \\
  \cmidrule(lr){1-3}
  & -1 &   \\
  &  & -I  \\
  \cmidrule(lr){1-3}
  &  & -I 
\end{bmatrix}, \quad
b = \begin{bmatrix} 0 \\ -g \\ 0 \\ 0 \\ 0\end{bmatrix},\quad
c = \begin{bmatrix} 1 \\ \lambda \\ 0 \end{bmatrix}, \quad
\mathcal{K} = \mathcal{Q}_{m+1} \times \mathcal{Q}_{n+1} \times \reals^n_+,
\end{align*}
with blank spaces representing zeros and the horizontal line denoting the
cone boundary. In this case, the parameters $F$, $g$ and $\lambda$
are just negated and copied into the problem data.

\paragraph{The canonicalization map.}
The full canonicalization procedure (which includes expanding graph implementations)
only runs \emph{the first time the problem is canonicalized}. When the same
problem is canonicalized in the future (\eg, with new parameter values), the
problem data $(A, b, c)$ can be obtained by multiplying a sparse matrix
representing $C$ by the parameter vector (and reshaping); the adjoint of
the derivative can be computed by just transposing the matrix. The na\"ive
alternative --- expanding graph implementations and extracting new problem data
every time parameters are updated (and differentiating through this algorithm
in the backward pass) --- is much slower (see \S\ref{sec:eval}).
The following lemma tells us that $C$ can be represented as a sparse matrix.
\begin{lemma}\label{lem:asa}
The canonicalizer map $C$ for a disciplined parametrized program can be represented with a
sparse matrix $Q \in \reals^{n \times p+1}$ and sparse tensor $R
\in \reals^{m \times n + 1 \times p+1}$, where $m$ is the dimension of the
constraints. Letting $\tilde \theta \in \reals^{p+1}$ denote
the concatenation of $\theta$ and the scalar offset $1$, the problem data
can be obtained as $c = Q \tilde \theta$ and $\begin{bmatrix}A &
b\end{bmatrix} = \sum_{i=1}^{p+1} R_{[:,:,i]}{\tilde \theta}_i$.
\end{lemma}
The proof is given in Appendix~\ref{apdx:canon}.

\subsection{Derivative of a conic solver}\label{sec:solver}
By applying the implicit function theorem
\citep{fiacco1968, dontchev2009implicit}
to the optimality conditions of a cone program,
it is possible to compute its derivative $\DD s(A, b, c)$.
To compute $\DD^T s(A, b, c)$, we
follow the methods presented in \citep{agrawal2019derivative}
and \citep[\S7.3]{amos2019differentiable}. Our calculations are given in
Appendix~\ref{apdx:derivative}.

If the cone program is not
differentiable at a solution, we compute a heuristic quantity, as is common
practice in automatic differentiation \citep[\S14]{griewank2008evaluating}.
In particular, at non-differentiable points, a linear system that arises in the
computation of the derivative might fail to be invertible. When this happens,
we compute a least-squares solution to the system instead. See
Appendix~\ref{apdx:derivative} for details.

\subsection{Solution retrieval}\label{sec:retrieval}
The cone program obtained by canonicalizing a DPP-compliant problem uses
the variable $\tilde{x} = (x, s) \in \reals^{n} \times \reals^{k}$, where $s \in \reals^k$ is a slack
variable. If $\tilde{x}^\star = (x^\star, s^\star)$ is optimal for the cone
program, then $x^\star$ is optimal for the original problem (up to reshaping
and scaling by a constant). As such, a solution to the original problem can be
obtained by slicing, \ie, $R(\tilde x^\star) = x^\star$. This map is evidently
linear.

\section{Implementation}\label{sec:impl}
We have implemented DPP and the reduction to ASA form in version 1.1 of CVXPY,
a Python-embedded DSL for convex optimization \citep{diamond2016cvxpy,
agrawal2018rewriting}; our implementation extends CVXCanon, an open-source
library that reduces affine expression trees to matrices \citep{miller2015}.
We have also implemented differentiable convex optimization layers in PyTorch
and TensorFlow 2.0. These layers implement the forward and backward maps
described in \S\ref{sec:derivative}; they also efficiently support batched inputs (see
\S\ref{sec:eval}).

We use the the \texttt{diffcp} package \citep{agrawal2019derivative} to obtain
derivatives of cone programs. We modified this package for performance:
we ported much of it from Python to C++, added an option to compute the
derivative using a dense direct solve, and made the forward and backward
passes amenable to parallelization.

Our implementation of DPP and ASA form, coupled with our PyTorch and TensorFlow layers,
makes our software the first DSL for differentiable convex optimization layers.
Our software is open-source. CVXPY and our layers are available at
\begin{center}
\url{https://www.cvxpy.org}, \quad \url{https://www.github.com/cvxgrp/cvxpylayers}.
\end{center}

\paragraph{Example.}
Below is an example of how to specify the problem (\ref{eqn:asa}) using
CVXPY 1.1.
\begin{lstlisting}
import cvxpy as cp

m, n = 20, 10
x = cp.Variable((n, 1))
F = cp.Parameter((m, n))
g = cp.Parameter((m, 1))
lambd = cp.Parameter((1, 1), nonneg=True)
objective_fn = cp.norm(F @ x - g) + lambd * cp.norm(x)
constraints = [x >= 0]
problem = cp.Problem(cp.Minimize(objective_fn), constraints)
assert problem.is_dpp()
\end{lstlisting}

The below code shows how to use our PyTorch layer to solve and backpropagate
through \texttt{problem} (the code for our TensorFlow layer is almost identical;
see Appendix~\ref{apdx:tf}).
\begin{lstlisting}
import torch
from cvxpylayers.torch import CvxpyLayer

F_t = torch.randn(m, n, requires_grad=True)
g_t = torch.randn(m, 1, requires_grad=True)
lambd_t = torch.rand(1, 1, requires_grad=True)
layer = CvxpyLayer(
    problem, parameters=[F, g, lambd], variables=[x])
x_star, = layer(F_t, g_t, lambd_t)
x_star.sum().backward()
\end{lstlisting}

Constructing \texttt{layer} in line \texttt{7-8} canonicalizes \texttt{problem}
to extract $C$ and $R$, as described in \S\ref{sec:canon}.
Calling \texttt{layer} in line \texttt{9} applies the map $R \circ s \circ C$
from \S\ref{sec:derivative}, returning a solution to the problem. Line \texttt{10}
computes the gradients of summing \texttt{x\_star}, with respect to \texttt{F\_t},
\texttt{g\_t}, and \texttt{lambd\_t}.

\section{Examples}\label{sec:ex}
In this section, we present two applications of differentiable convex
optimization, meant to be suggestive of possible use cases for our layer. We
give more examples in Appendix~\ref{apdx:more-ex}.

\begin{figure}
\centering
\adjustbox{valign=t}{\begin{minipage}[b]{0.49\textwidth}
\includegraphics[width=\textwidth]{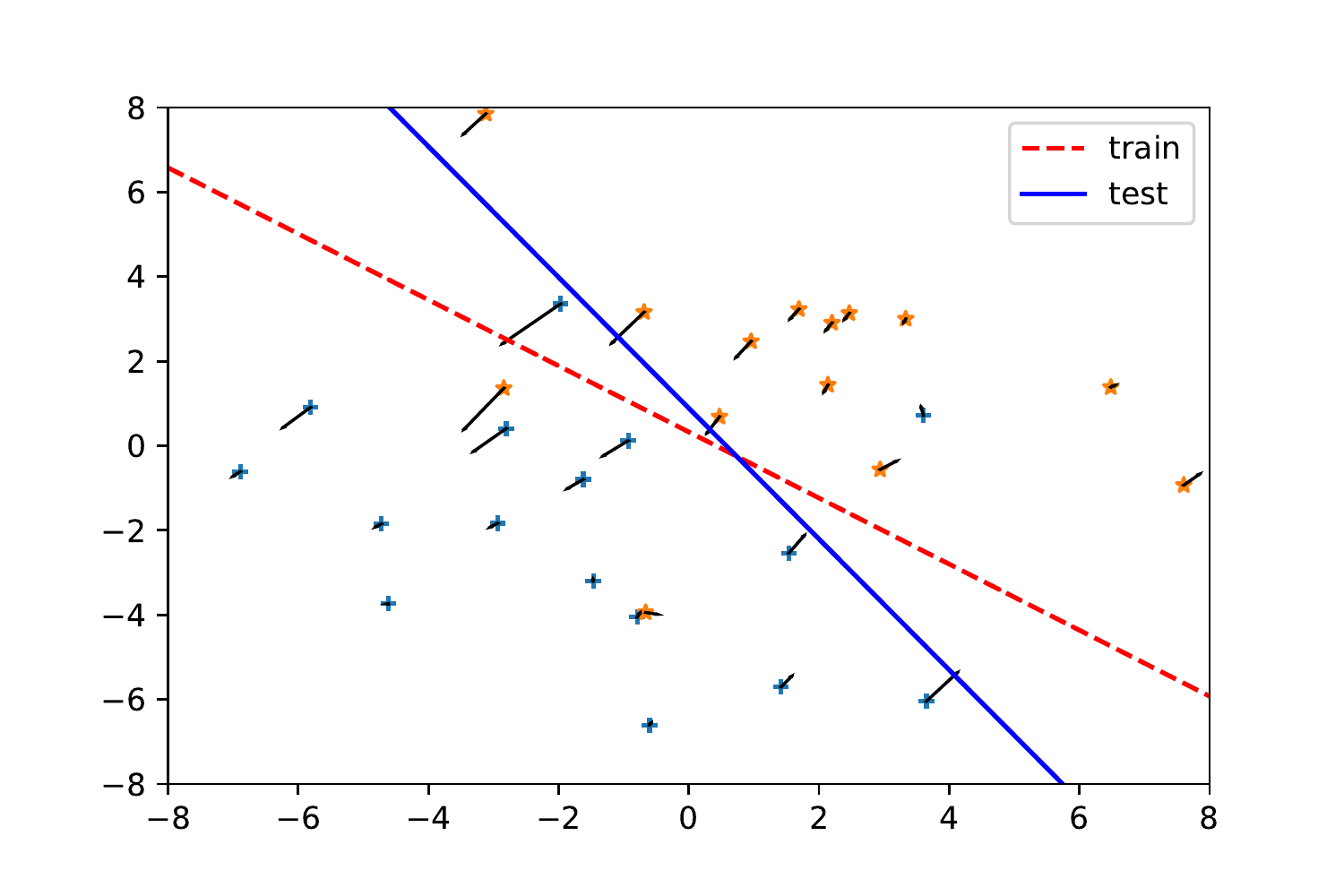}
\caption{Gradients (black lines) of the logistic test loss
with respect to the training data.}
\label{fig:logreg}
\end{minipage}}%
\hfill
\adjustbox{valign=t}{\begin{minipage}[c]{0.49\textwidth}
  \includegraphics[width=\textwidth]{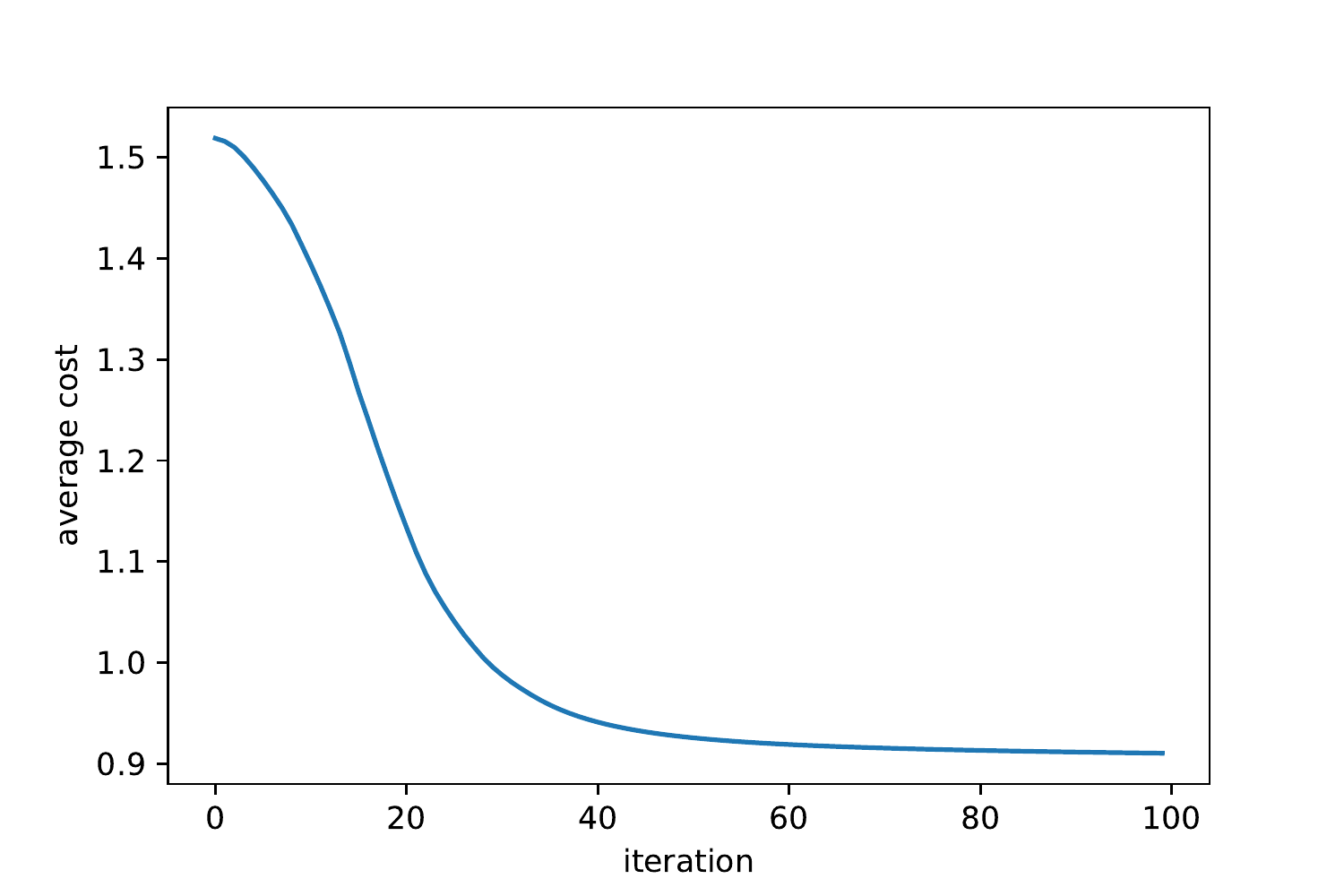}
\caption{Per-iteration cost while learning an ADP policy for stochastic control.}
\label{fig:control}
\end{minipage}}
\end{figure}

\subsection{Data poisoning attack}
We are given training data $(x_i, y_i)_{i=1}^{N}$,
where $x_i\in\reals^n$ are feature vectors and $y_i\in\{0,1\}$ are the labels.
Suppose we fit a model for this classification problem by solving
\begin{equation}
\begin{array}{ll}
\mbox{minimize} & \frac{1}{N}\sum_{i=1}^N \ell(\theta; x_i, y_i) + r(\theta),
\end{array}
\label{eq:trainlinear}
\end{equation}
where the loss function $\ell(\theta; x_i, y_i)$ is convex in $\theta \in \reals^n$ and $r(\theta)$ is a convex
regularizer. We hope that the test loss $\mathcal{L}^{\mathrm{test}}(\theta) =
\frac{1}{M}\sum_{i=1}^M \ell(\theta; \tilde x_i, \tilde y_i)$ is small, where
$(\tilde x_i, \tilde y_i)_{i=1}^{M}$ is our test set.

Assume that our training data is subject to a data poisoning attack
\citep{biggio2018wild,jagielski2018manipulating},
before it is supplied to us. The adversary has full knowledge of our modeling
choice, meaning that they know the form of \eqref{eq:trainlinear}, and seeks
to perturb the data to maximally increase our loss on the test
set, to which they also have access. The adversary is permitted to apply an
additive perturbation $\delta_i \in \reals^n$ to each of the training points $x_i$,
with the perturbations satisfying $\norm{\delta_i}_\infty \leq 0.01$.

Let $\theta^\star$ be optimal for \eqref{eq:trainlinear}.
The gradient of
the test loss with respect to a training data point,
$\nabla_{x_i} \mathcal{L}^{\mathrm{test}}(\theta^\star))$.gives the direction in which the
point should be moved to achieve the greatest
increase in test loss. Hence, one reasonable adversarial policy is to set $x_i
\coloneqq x_i +
(.01)\mathbf{sign}(\nabla_{x_i}\mathcal{L}^{\mathrm{test}}(\theta^\star))$. The
quantity $(0.01)\sum_{i=1}^N \norm{\nabla_{x_i}
\mathcal{L}^{\mathrm{test}}(\theta^\star)}_1$ is the predicted increase in
our test loss due to the poisoning.

\paragraph{Numerical example.}
We consider 30 training points and 30 test points in
$\reals^2$, and we fit a logistic model with elastic-net regularization. This
problem can be written using DPP, with $x_i$ as parameters
(see Appendix~\ref{apdx:logreg} for the code). We used our convex optimization
layer to fit this model and obtain the gradient of the test loss with respect
to the
training data. Figure~\ref{fig:logreg} visualizes the results. The orange ($\star$) and
blue (+) points are training data, belonging to different classes. The red
line (dashed) is the hyperplane learned by
fitting the the model, while the blue line (solid) is the
hyperplane that minimizes the test loss. The gradients are visualized as black
lines, attached to the data points. Moving the points in the gradient
directions torques the learned hyperplane away from the optimal hyperplane for
the test set.

\subsection{Convex approximate dynamic programming}

We consider a stochastic control problem of the form
\begin{equation}
\begin{array}{ll}
\mbox{minimize} & \underset{T \to \infty}\lim {\mathbb E} \left[\frac{1}{T} \sum_{t=0}^{T-1} \norm{x_t}_2^2 + \norm{\phi(x_t)}_2^2\right]\\[.2cm]
\mbox{subject to} & x_{t+1} = Ax_t + B\phi(x_t) + \omega_t, \quad t=0,1,\ldots,
\end{array}
\label{eq:adp}
\end{equation}
where $x_t\in\reals^n$ is the state, $\phi:\reals^n \to \mathcal U \subseteq \reals^m$ is
the policy, $\mathcal U$ is a convex set representing the allowed set of controls,
and $\omega_t\in\Omega$ is a (random, i.i.d.) disturbance.
Here the variable is the policy $\phi$, and the expectation is taken over
disturbances and the initial state $x_0$. If $\mathcal U$ is not an affine
set, then this problem is in general very difficult to solve
\cite{kalman1964linear, barratt2018stochastic}.

\paragraph{ADP policy.}
A common heuristic for solving \eqref{eq:adp} is
approximate dynamic programming (ADP), which parametrizes $\phi$
and replaces the minimization over functions $\phi$ with a minimization over parameters.
In this example, we take $\mathcal U$ to be the unit ball and we represent $\phi$
as a quadratic \emph{control-Lyapunov} policy \cite{wang2010fast}.
Evaluating $\phi$ corresponds to solving the SOCP
\begin{equation}
\begin{array}{ll}
\mbox{minimize} & u^T P u + x_t^T Q u + q^T u \\
\mbox{subject to} & \norm{u}_2 \leq 1,
\end{array}
\label{eq:policy}
\end{equation}
with variable $u$ and parameters $P$, $Q$, $q$, and $x_t$. We can run
stochastic gradient descent (SGD) on $P$, $Q$, and $q$ to
approximately solve \eqref{eq:adp}, which requires differentiating
through \eqref{eq:policy}. Note that if $u$ were unconstrained, \eqref{eq:adp}
could be solved exactly, via linear quadratic regulator (LQR) theory
\citep{kalman1964linear}.
The policy \eqref{eq:policy} can be
written using DPP (see Appendix~\ref{apdx:logreg} for the code).

\paragraph{Numerical example.}
Figure~\ref{fig:control} plots the estimated average cost for each iteration of
gradient descent for a numerical example, with $x \in
\reals^2$ and $u \in \reals^3$, a time horizon of $T=25$, and a batch
size of $8$. We initialize our policy's parameters with the LQR solution,
ignoring the constraint on $u$. This method decreased the average cost
by roughly 40\%.

\section{Evaluation}\label{sec:eval}

Our implementation substantially lowers the barrier to using convex
optimization layers. Here, we show that our implementation substantially
reduces canonicalization time. Additionally, for dense problems, our
implementation is competitive (in execution time) with a specialized solver for
QPs; for sparse problems, our implementation is much faster.

\paragraph{Canonicalization.}

\begin{table}[t]
  \caption{Time (ms) to canonicalize examples,
  across 10 runs.}\label{tab:canon}
  \centering
  \begin{tabular}{lll}
    \toprule
         & Logistic regression     & Stochastic control \\
    \midrule
    CVXPY 1.0.23         & 18.9 $\pm$ 1.75 & 12.5 $\pm$ 0.72 \\
    CVXPY 1.1 & 1.49 $\pm$ 0.02 & 1.39 $\pm$ 0.02 \\
    \bottomrule
  \end{tabular}
\end{table}

Table~\ref{tab:canon} reports the
time it takes to canonicalize the logistic regression and stochastic
control problems from \S\ref{sec:ex}, comparing CVXPY version 1.0.23 with
CVXPY 1.1.
Each canonicalization was performed on a single core of an
unloaded Intel i7-8700K processor.
We report the
average time and standard deviation across 10 runs, excluding a warm-up run.
Our extension achieves on average an order-of-magnitude speed-up since
computing $C$ via a sparse matrix multiply is much more efficient than
going through the DSL.

\paragraph{Comparison to specialized layers.}

\begin{figure}[t]
  \centering
  \begin{subfigure}{.49\textwidth}
    \centering
    \includegraphics[height=40mm]{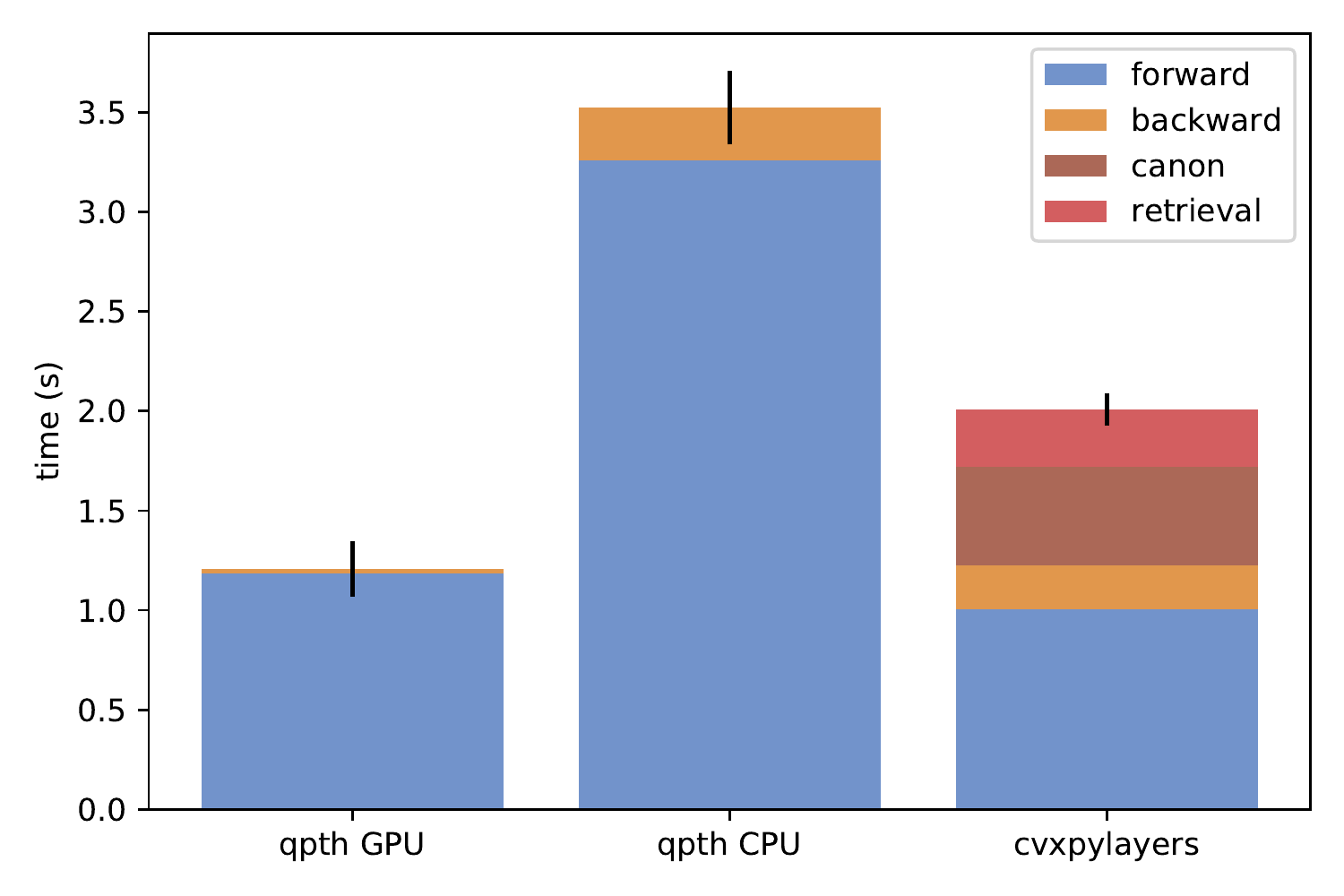}
    \caption{Dense QP, batch size of 128.}
    \label{fig:dense}
  \end{subfigure}
  \begin{subfigure}{.49\textwidth}
    \centering
    \includegraphics[height=40mm]{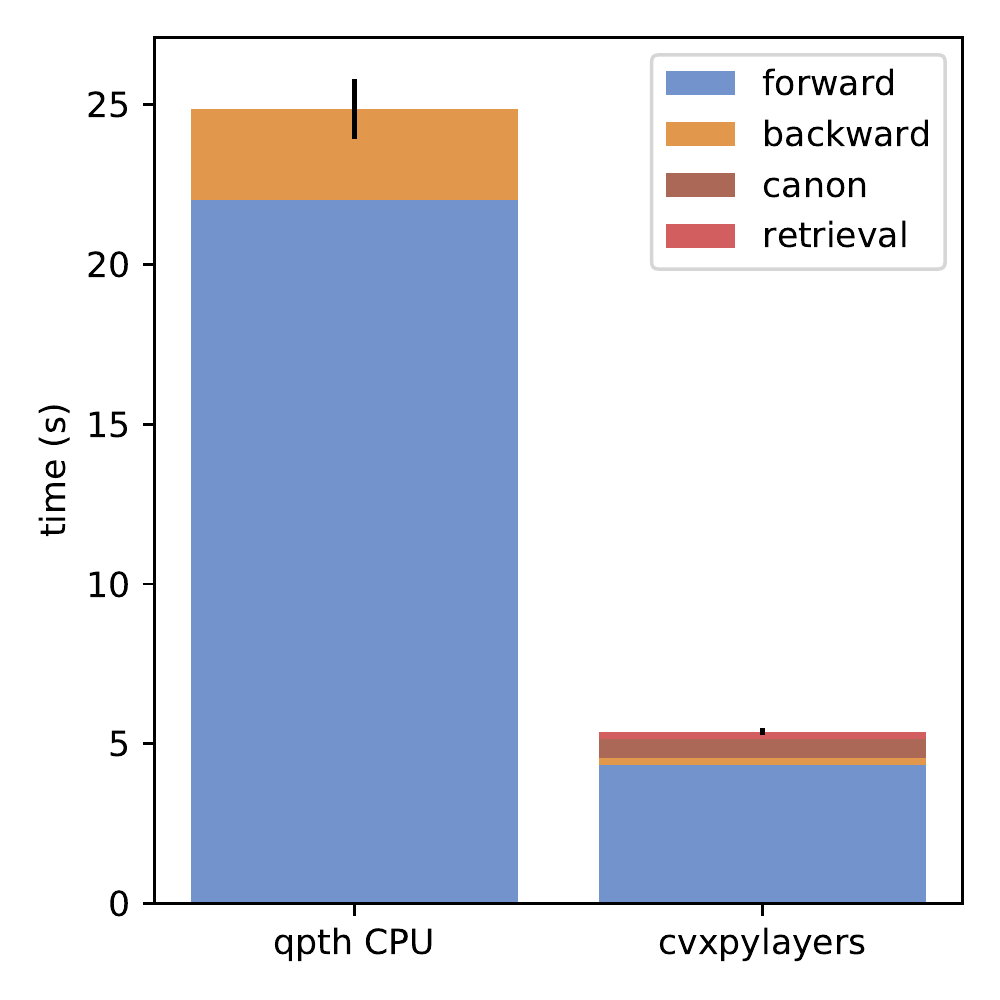}
    \caption{Sparse QP, batch size of 32.}
    \label{fig:sparse}
  \end{subfigure}
  \caption{Comparison of our PyTorch \texttt{CvxpyLayer} to \qpth,
    over 10 trials. For \texttt{cvxpylayers}, we separate out the canonicalization and solution
    retrieval times, to allow for a fair comparison.}
  \label{fig:comparison}
\end{figure}

We have implemented a batched solver and backward pass
for our differentiable CVXPY layer that makes it competitive with
the batched QP layer \qpth from \citep{amos2017optnet}.
\cref{fig:comparison} compares the runtimes of our PyTorch
\texttt{CvxpyLayer} and \qpth on a dense and sparse QP.
The sparse problem is too large for \qpth to run in
GPU mode.
The QPs have the form
\begin{equation}
\begin{array}{ll}
\mbox{minimize} & \frac{1}{2} x^T Q x + p^T x \\ [.1cm]
\mbox{subject to} & Ax=b,\\
& Gx \leq h,
\end{array}
\label{eq:qp}
\end{equation}
with variable $x\in\reals^n$, and problem data
$Q\in\reals^{n \times n}$, $p\in\reals^n$, $A\in\reals^{m\times n}$,
$b\in\reals^m$, $G\in\reals^{p \times n}$, and $h\in\reals^p$.
The dense QP has $n=128$, $m=0$, and $p=128$.
The sparse QP has $n=1024$, $m=1024$, and $p=1024$
and $Q$, $A$, and $G$ each have 1\% nonzeros
(See Appendix~\ref{apdx:more-ex} for the code).
We ran this experiment on a machine with a
6-core Intel i7-8700K CPU,
32 GB of memory, and an Nvidia GeForce 1080 TI GPU with 11 GB of memory.

Our implementation is competitive with \qpth for the dense QP,
even on the GPU, and roughly 5 times faster for the sparse QP.
Our backward pass for the dense QP uses our extension to \texttt{diffcp};
we explicitly materialize the derivatives of the
cone projections and use a direct solve. Our backward pass for the sparse QP
uses sparse operations and LSQR \citep{paige1982lsqr}, significantly
outperforming \qpth (which cannot exploit sparsity). Our layer runs on the CPU,
and implements batching via Python multi-threading, with a parallel for loop
over the examples in the batch for both the forward and backward passes.
We used 12 threads for our experiments.

\section{Discussion}

\paragraph{Other solvers.}
Solvers that are specialized to subclasses of convex programs are often faster
than more general conic solvers. For example, one might use OSQP
\citep{stellato2017osqp} to solve QPs, or gradient-based methods like L-BFGS
\citep{liu1989limited} or SAGA \citep{defazio2014saga} for empirical risk
minimization. Because CVXPY lets developers add specialized solvers as
additional back-ends, our implementation of DPP and ASA form can be easily extended to
other problem classes. We plan to interface QP solvers in future work.

\paragraph{Nonconvex problems.} It is possible to differentiate through
nonconvex problems, either analytically \citep{fiacco1983nlp,
pirnay2012optimal, amos2018differentiable} or by unrolling SGD
\citep{domke2012generic,belanger2017end,metz2016unrolled,goodfellow2013multi,stoyanov2011empirical,brakel2013training,finn2017model},
Because convex programs can typically be solved efficiently and to high
accuracy, it is preferable to use convex optimization layers over nonconvex
optimization layers when possible. This is especially true in the setting of
low-latency inference. The
use of differentiable nonconvex programs in end-to-end learning pipelines,
discussed in \citep{gould2019deep}, is an interesting direction for future
research.

\clearpage
\section*{Acknowledgments} We gratefully acknowledge discussions with Eric Chu,
who designed and implemented a code generator for SOCPs
\citep{chu2013code, chu2017qcml}, Nicholas Moehle, who designed and
implemented a basic version of a code generator for convex optimization in
unpublished work, and Brendan O'Donoghue.
We also would like to thank the anonymous reviewers, who provided us
with useful suggestions that improved the paper.
S. Barratt is supported by the National Science Foundation Graduate Research Fellowship
under Grant No. DGE-1656518.

\printbibliography{}

\clearpage
\appendix
\section{The canonicalization map}\label{apdx:canon}
In this appendix, we provide a proof of Lemma~\ref{lem:asa}. 
We compute $Q$ and $R$ via a reduction on the affine expression trees
that represent the canonicalized problem.
Let $f$ be the root node with arguments (descendants) $g_1,
\ldots, g_n$. Then we obtain tensors $T_1,\ldots,T_n$ representing the (linear)
action of $f$ on each argument. We recurse on each subtree $g_i$ and obtain
tensors $S_1,\ldots,S_n$. Due to the DPP rules, for $i=1,\ldots,n$, we either
have $(T_i)_{j,k,\ell} = 0$ for $\ell \neq p+1$ or $(S_i)_{j,k,\ell} = 0$ for
$\ell \neq p+1$. We define an operation $\psi(T_i,S_i)$ such that in the first
case, $\psi(T_i, S_i) =
\sum_{\ell=1}^{p+1}(T_{i})_{[:,:,p+1]}(S_{i})_{[:,:,\ell]}$, and in the second
case
$\psi(T_i, S_i) = \sum_{\ell=1}^{p+1}(T_{i})_{[:,:,\ell]}(S_{i})_{[:,:,p+1]}$.
The tree rooted at $f$ then evaluates to
$S_0 = \psi(T_1,S_1) + \cdots + \psi(T_n,S_n)$.

The base case of the recursion corresponds to the tensors produced when a
variable, parameter, or constant node is evaluated. (These are the leaf nodes
of an affine expression tree.)
\begin{itemize}
\item A variable leaf $x \in \reals^d$ produces a tensor $T \in
\reals^{d \times n+1 \times 1}$, where $T_{i,j,1}=1$ if $i$ maps to $j$ in the
vector containing all variables, 0 otherwise.
\item A parameter leaf $p \in \reals^d$ produces a tensor $T \in
\reals^{d \times 1 \times p+1}$, where $T_{i,1,j}=1$ if $i$ maps to $j$ in the
vector containing all parameters, 0 otherwise.
\item A constant leaf $c \in \reals^d$ produces a tensor $T \in
\reals^{d \times 1 \times 1}$, where $T_{i,1,1}=c_i$ for $i=1,\ldots,d$.
\end{itemize}

\section{Derivative of a cone program}\label{apdx:derivative}
In this appendix, we show how to differentiate through a cone program.
We first present some preliminaries.

\paragraph{Primal-dual form of a cone program.}

A (convex) cone program is given by
\begin{center}
\begin{tabular}{p{6.5cm}p{6.5cm}}
{\begin{equation*}
      \begin{array}{lll}
        \text{(P)}
        &\text{minimize}  &c^T x\\
        &\text{subject to} &  Ax+s=b\\
        &  &s\in  \mathcal{K},
      \end{array}
 \end{equation*}}
  &
  { \vspace{-.8em}\begin{equation}
      \label{eq:pdcp}
      \begin{array}{lll}
        \text{(D)}&\text{minimize}& b^T y\\
        &\text{subject to}& A^T y+c=0\\
        &&y\in  \mathcal{K}^*.
      \end{array}
  \end{equation}}
\end{tabular}
\end{center}
Here $x\in \reals^n$ is the \emph{primal} variable, $y\in \reals^m$ is the
\emph{dual} variable, and $s\in \reals^m$ is the primal \emph{slack} variable.
The set $\mathcal{K}\subseteq \reals^m$ is a nonempty, closed, convex cone with
\emph{dual cone} $\mathcal{K}^*\subseteq \reals^m$.
We call  $(x,y,s)$
a solution of the primal-dual
cone program \eqref{eq:pdcp}
if it satisfies the KKT conditions:
\[
Ax +s =b, \quad
A^T y + c = 0, \quad
s \in \mathcal{K}, \quad
y \in \mathcal{K}^*, \quad
s^T y = 0.
\]

Every convex optimization problem can be reformulated as a convex cone program.

\paragraph{Homogeneous self-dual embedding.}
The homogeneous self-dual
embedding reduces the process of solving \eqref{eq:pdcp}
to finding a zero of a certain residual map \cite{ye1994nl}.
Letting $N=n+m+1$, the embedding
uses the variable $z\in \reals^N$, which we partition as
$(u,v,w) \in \reals^n \times \reals^m \times \reals$.
The \emph{normalized residual map} introduced in \cite{busseti2018solution} is
the function
$ \mathcal{N}: \reals^N
\times \reals^{N \times N}
\to \reals^N$,
defined by
\[
\mathcal{N}(z,Q)
=\big((Q-I)\Pi+I\big)(z/|w|),
\]
where $\Pi$ denotes projection onto $\reals^n \times \mathcal {K}^* \times \reals_+$,
and $Q$ is the skew-symmetric matrix
\BEQ
  Q = \begin{bmatrix}
    0 & A^T & c\\
    -A & 0 & b \\
    -c^T & -b^T & 0
  \end{bmatrix}.
  \label{eq:skew}
\EEQ
If $\mathcal{N}(z,Q)=0$ and $w>0$,
we can use $z$ to construct a solution
of the primal-dual pair \eqref{eq:pdcp} as
\BEQ
(x,y,s) = (u, \Pi_\mathcal{K^*}(v),
\Pi_{\mathcal{K}^*}(v)-v)/w,
\label{eq:sol}
\EEQ
where $\Pi_{\mathcal K^*}(v)$ denotes the projection of $v$
onto $\mathcal K^*$.
From here onward, we assume that $w=1$.
(If this is not the case, we can scale $z$ such that it is the case.)

\paragraph{Differentiation.}
A conic solver is a numerical algorithm for solving
\eqref{eq:pdcp}.
We can view a conic solver as a function
$\psi:
\reals^{m\times n} \times
\reals^m\times \reals^n \to \reals^{n+2m}$
mapping the problem data
$(A,b,c)$ to a solution $(x,y,s)$.
(We assume that the cone $\mathcal K$ is fixed.)
In this section we derive expressions for the derivative of $\psi$,
assuming that $\mathcal S$ is in fact differentiable.
Interlaced with our derivations, we describe how to numerically evaluate
the adjoint of the derivative map, which is necessary for backpropagation.

Following \cite{agrawal2019derivative} and \citep[Section~7]{amos2019differentiable},
we can express $\psi$ as the composition
$\phi \circ s \circ Q$, where
\begin{itemize}
  \item $Q: \reals^{m \times n} \times \reals^m \times \reals^n \to \mathcal \reals^{N \times N}$
    maps the problem data to $Q$, given
    by \eqref{eq:skew},
  \item $s: \reals^{N \times N} \to \reals^N$ solves the homogeneous
    self-dual embedding, which we can implicitly differentiate, and
  \item $\phi: \reals^N \to \reals^n \times \reals^m \times \reals^m$
    maps $z$ to the primal-dual pair, given by \eqref{eq:sol}.
\end{itemize}

To backpropagate through $\psi$, we need to compute
the adjoint of the derivative of $\psi$ at $(A,b,c)$ applied to the vector
$(\mathsf{d}x, \mathsf{d}y, \mathsf{d}s)$, or
\[
(\mathsf{d}A, \mathsf{d}b, \mathsf{d}c) = \DD^T\psi(A,b,c)(\mathsf{d}x, \mathsf{d}y, \mathsf{d}s) = \DD^TQ(A, b, c) \DD^Ts(Q) \DD^T \phi(z) (\mathsf{d}x, \mathsf{d}y, \mathsf{d}s).
\]
Since our layer only outputs the primal solution $x$, we can simplify the
calculation by taking $\mathsf{d}y = \mathsf{d}s = 0$. By \eqref{eq:sol},
\[
\mathsf{d}z = \DD^T \phi(z) (\mathsf{d}x, 0, 0) =
\begin{bmatrix}
\mathsf{d}x \\
0 \\
-x^T\mathsf{d}x
\end{bmatrix}.
\]
We can compute $\mathsf{D}s(Q)$ by implicitly differentiating the normalized
residual map:
\BEQ
\label{eq:implicit}
\mathsf{D}s(Q) = -(\DD_z \mathcal{N}(s(Q), Q))^{-1} \DD_Q \mathcal{N}(s(Q), Q).
\EEQ
This gives
\[
\mathsf{d}Q = \mathsf{D}^Ts(Q)\mathsf{d}z = -(M^{-T}\mathsf{d}z) \Pi(z)^T,
\]
where $M = (Q - I)\DD\Pi(z) + I$. Computing $g = M^{-T}\mathsf{d}z$ via
a direct method (\ie, materializing $M$, factorizing it, and back-solving)
can be impractical when $M$ is large. Instead, one might use a Krylov method
like LSQR \citep{paige1982lsqr} to solve
\begin{equation}
\begin{array}{ll}
\underset{g}{\mbox{minimize}} & \|M^Tg - \mathsf{d}z\|_2^2,
\end{array}\label{eq:M}
\end{equation}
which only requires multiplication by $M$ and $M^T$. Instead of computing
$\mathsf{d}Q$ as an outer product, we only obtain its nonzero entries. Finally,
partitioning $\mathsf{d}Q$ as
\[
\mathsf{d}Q = \begin{bmatrix}
\mathsf{d}Q_{11} & \mathsf{d}Q_{12} & \mathsf{d}Q_{13} \\
\mathsf{d}Q_{21} & \mathsf{d}Q_{22} & \mathsf{d}Q_{23} \\
\mathsf{d}Q_{31} & \mathsf{d}Q_{32} & \mathsf{d}Q_{33} \\
\end{bmatrix},
\]
we obtain
\BEAS
\mathsf{d}A &=& -\mathsf{d}Q_{12}^T + \mathsf{d}Q_{21} \\
\mathsf{d}b &=& -\mathsf{d}Q_{23} + \mathsf{d}Q_{32}^T \\
\mathsf{d}c &=& -\mathsf{d}Q_{13} + \mathsf{d}Q_{31}^T.
\EEAS
\paragraph{Non-differentiability.}
To implicitly differentiate the solution map in $\eqref{eq:implicit}$, we assumed
that the $M$ was invertible. When $M$ is not
invertible, we approximate $\mathsf{d}Q$ as $-g^\mathrm{ls} \Pi(z)^T$, where
$g^\mathrm{ls}$ is a least-squares solution to $\eqref{eq:M}$.

\newpage
\section{Examples}\label{apdx:ex}
This appendix includes code for the examples presented in \S\ref{sec:ex}.

\paragraph{Logistic regression.}\label{apdx:logreg}
The code for the logistic regression problem is below:
\begin{lstlisting}
import cvxpy as cp
from cvxpylayers.torch import CvxpyLayer

beta = cp.Variable((n, 1))
b = cp.Variable((1, 1))
X = cp.Parameter((N, n))

log_likelihood = (1. / N) * cp.sum(
    cp.multiply(Y, X @ beta + b) - cp.logistic(X @ beta + b)
)
regularization = -0.1 * cp.norm(beta, 1) -0.1 * cp.sum_squares(beta)

prob = cp.Problem(cp.Maximize(log_likelihood + regularization))
fit_logreg = CvxpyLayer(prob, parameters=[X], variables=[beta, b])
\end{lstlisting}

\paragraph{Stochastic control.}\label{apdx:stoch}
The code for the stochastic control problem \eqref{eq:adp} is below:
\begin{lstlisting}
import cvxpy as cp
from cvxpylayers.torch import CvxpyLayer

x_cvxpy = cp.Parameter((n, 1))
P_sqrt_cvxpy = cp.Parameter((m, m))
P_21_cvxpy = cp.Parameter((n, m))
q_cvxpy = cp.Parameter((m, 1))

u_cvxpy = cp.Variable((m, 1))
y_cvxpy = cp.Variable((n, 1))

objective = .5 * cp.sum_squares(P_sqrt_cvxpy @ u_cvxpy) + x_cvxpy.T @ y_cvxpy + q_cvxpy.T @ u_cvxpy
prob = cp.Problem(cp.Minimize(objective),
  [cp.norm(u_cvxpy) <= .5, y_cvxpy == P_21_cvxpy @ u_cvxpy])

policy = CvxpyLayer(prob,
  parameters=[x_cvxpy, P_sqrt_cvxpy, P_21_cvxpy, q_cvxpy],
  variables=[u_cvxpy])

\end{lstlisting}

\section{TensorFlow layer}\label{apdx:tf}
In \S\ref{sec:impl}, we showed how to implement the problem \eqref{eqn:asa}
using our PyTorch layer. The below code shows how to implement the same problem
using our TensorFlow 2.0 layer.
\begin{lstlisting}
import tensorflow as tf
from cvxpylayers.tensorflow import CvxpyLayer

F_t = tf.Variable(tf.random.normal(F.shape))
g_t = tf.Variable(tf.random.normal(g.shape))
lambd_t = tf.Variable(tf.random.normal(lambd.shape))
layer = CvxpyLayer(problem, parameters=[F, g, lambd], variables=[x])
with tf.GradientTape() as tape:
  x_star, = layer(F_t, g_t, lambd_t)
dF, dg, dlambd = tape.gradient(x_star, [F_t, g_t, lambd_t])
\end{lstlisting}

\section{Additional examples}\label{apdx:more-ex}
In this appendix we provide additional examples of constructing
differentiable convex optimization layers using our implementation.
We present the implementation of common neural networks layers,
even though analytic solutions exist for some of these operations.
These layers can be modified in simple ways such that they
do \emph{not} have analytical solutions. In the below problems, the
optimization variable is $y$ (unless stated otherwise). We also show how prior
work on differentiable convex optimization layers such as OptNet
\cite{amos2017optnet} is captured by our framework.

The \textbf{ReLU}, defined by $f(x) = \max\{0, x\}$,
can be interpreted as projecting a point $x\in\reals^n$ onto
the non-negative orthant as
\begin{equation*}
\begin{array}{ll}
  \mbox{minimize} & \frac{1}{2}||x-y||_2^2 \\
  \mbox{subject to} & y \geq 0.
\end{array}
\end{equation*}
We can implement this layer with:
\begin{lstlisting}
x = cp.Parameter(n)
y = cp.Variable(n)
obj = cp.Minimize(cp.sum_squares(x-y))
cons = [y >= 0]
prob = cp.Problem(obj, cons)
layer = CvxpyLayer(prob, parameters=[x], variables=[y])
\end{lstlisting}

The \textbf{sigmoid} or \textbf{logistic} function, defined by $f(x) =
(1+e^{-x})^{-1}$, can be interpreted as projecting a point $x\in\reals^n$ onto
the interior of the unit hypercube as
\begin{equation*}
\begin{array}{ll}
  \mbox{minimize} & -x^\top y -H_b(y) \\
  \mbox{subject to} & 0 < y < 1,
\end{array}
\end{equation*}
where $H_b(y) = - \left(\sum_i y_i\log y_i + (1-y_i)\log (1-y_i)\right)$ is the
binary entropy function.
This is proved, \eg, in \cite[Section 2.4]{amos2019differentiable}.
We can implement this layer with:
\begin{lstlisting}
x = cp.Parameter(n)
y = cp.Variable(n)
obj = cp.Minimize(-x.T*y - cp.sum(cp.entr(y) + cp.entr(1.-y)))
prob = cp.Problem(obj)
layer = CvxpyLayer(prob, parameters=[x], variables=[y])
\end{lstlisting}

The \textbf{softmax}, defined by $f(x)_j = e^{x_j} / \sum_i e^{x_i}$,
can be interpreted as projecting a point $x\in\reals^n$ onto
the interior of the $(n-1)$-simplex
$\Delta_{n-1}=\{p\in\reals^n\; \vert\; 1^\top p = 1 \; \; {\rm and} \;\; p \geq 0 \}$
as
\begin{equation*}
\begin{array}{ll}
\mbox{minimize} & -x^\top y - H(y) \\
\mbox{subject to} & 0 < y < 1, \\
& 1^\top y = 1,
\end{array}
\end{equation*}
where $H(y) = -\sum_i y_i \log y_i$ is the entropy function.
This is proved, \eg, in \cite[Section 2.4]{amos2019differentiable}.
We can implement this layer with:
\begin{lstlisting}
x = cp.Parameter(d)
y = cp.Variable(d)
obj = cp.Minimize(-x.T*y - cp.sum(cp.entr(y)))
cons = [sum(y) == 1.]
prob = cp.Problem(obj, cons)
layer = CvxpyLayer(prob, parameters=[x], variables=[y])
\end{lstlisting}

\newpage
The \textbf{sparsemax} \cite{martins2016softmax} does a Euclidean
projection onto the simplex as
\begin{equation*}
\begin{array}{ll}
\mbox{minimize} & ||x-y||_2^2 \\
\mbox{subject to} & 1^\top y = 1, \\
& 0 \leq y \leq 1.
\end{array}
\end{equation*}
We can implement this layer with:
\begin{lstlisting}
x = cp.Parameter(n)
y = cp.Variable(n)
obj = cp.sum_squares(x-y)
cons = [cp.sum(y) == 1, 0. <= y, y <= 1.]
prob = cp.Problem(cp.Minimize(obj), cons)
layer = CvxpyLayer(prob, [x], [y])
\end{lstlisting}

The \textbf{constrained softmax} \cite{martins2017learning} solves
the optimization problem
\begin{equation*}
\begin{array}{ll}
  \mbox{minimize} & -x^\top y - H(y) \\
  \mbox{subject to} & 1^\top y = 1, \\ 
  & y \leq u, \\
  & 0 < y < 1.
\end{array}
\end{equation*}
We can implement this layer with:
\begin{lstlisting}
x = cp.Parameter(n)
y = cp.Variable(n)
obj = -x*y-cp.sum(cp.entr(y))
cons = [cp.sum(y) == 1., y <= u]
prob = cp.Problem(cp.Minimize(obj), cons)
layer = CvxpyLayer(prob, [x], [y])
\end{lstlisting}

The \textbf{constrained sparsemax} \cite{malaviya2018sparse} solves
the optimization problem
\begin{equation*}
\begin{array}{ll}
  \mbox{minimize} & ||x-y||_2^2, \\
  \mbox{subject to} & 1^\top y = 1, \\
  & 0 \leq y \leq u.
\end{array}
\end{equation*}
We can implement this layer with:
\begin{lstlisting}
x = cp.Parameter(n)
y = cp.Variable(n)
obj = cp.sum_squares(x-y)
cons = [cp.sum(y) == 1., 0. <= y, y <= u]
prob = cp.Problem(cp.Minimize(obj), cons)
layer = CvxpyLayer(prob, [x], [y])
\end{lstlisting}

The \textbf{Limited Multi-Label (LML)} layer \cite{amos2019lml}
solves the optimization problem
\begin{equation*}
\begin{array}{ll}
  \mbox{minimize} & -x^\top y - H_b(y) \\
  \mbox{subject to} & 1^\top y = k, \\
  & 0 < y < 1.
\end{array}
\end{equation*}
We can implement this layer with:
\begin{lstlisting}
x = cp.Parameter(n)
y = cp.Variable(n)
obj = -x*y-cp.sum(cp.entr(y))-cp.sum(cp.entr(1.-y))
cons = [cp.sum(y) == k]
prob = cp.Problem(cp.Minimize(obj), cons)
layer = CvxpyLayer(prob, [x], [y])
\end{lstlisting}

\paragraph{The OptNet QP.}
We can re-implement the OptNet QP layer \citep{amos2017optnet} in a few lines
of code. The OptNet layer is a solution to a convex quadratic program of the
form
\begin{equation*}
\begin{array}{ll}
\mbox{minimize} & \frac{1} {2}x^\top Qx + q^\top x \\
\mbox{subject to} & Ax = b, \\
                  & Gx \leq h,
\end{array}
\end{equation*}
where $x \in \reals^n$ is the optimization variable,
and the problem data are $Q \in \reals^{n \times n}$
(which is positive semidefinite),
$q \in \reals^n$,
$A\in \reals^{m \times n}$,
$b \in \reals^m$,
$G \in \reals^ {p \times n}$, and
$h \in \reals^{p}$.
We can implement this with:

\begin{lstlisting}
Q_sqrt = cp.Parameter((n, n))
q = cp.Parameter(n)
A = cp.Parameter((m, n))
b = cp.Parameter(m)
G = cp.Parameter((p, n))
h = cp.Parameter(p)
x = cp.Variable(n)
obj = cp.Minimize(0.5*cp.sum_squares(Q_sqrt*x) + q.T @ x)
cons = [A @ x == b, G @ x <= h]
prob = cp.Problem(obj, cons)
layer = CvxpyLayer(prob, parameters=[Q_sqrt, q, A, b, G, h], variables=[x])
\end{lstlisting}
Note that we take the matrix square-root of $Q$ in PyTorch, outside the
CVXPY layer, to get the derivative with respect to $Q$. DPP does not allow the
quadratic form atom to be parametrized, as discussed in \S\ref{sec:dpp}.

\end{document}